\documentclass[letterpaper, 10 pt, conference]{ieeeconf}  

\IEEEoverridecommandlockouts                              

\usepackage{amsmath} 

\usepackage{hyperref}
\hypersetup{
    colorlinks=true,
    linkcolor=blue,
    filecolor=magenta,      
    urlcolor=blue,
}

\usepackage{upgreek}
\usepackage{amsfonts,amssymb}
\usepackage{bm}
\usepackage{bbm}

\usepackage{amsthm} 
\usepackage{thmtools}
\usepackage{mathtools}
\usepackage{xspace}
\usepackage[font=footnotesize]{subcaption}
\usepackage{units}
\usepackage{booktabs} 
\usepackage{tabularx}
\usepackage{tabulary}
\usepackage[usenames,dvipsnames]{xcolor} 
\usepackage{diagbox}
\usepackage[ruled,vlined,linesnumbered,algo2e]{algorithm2e}  
\SetKwComment{Comment}{$\triangleright$\ }{}
\usepackage{ifthen,version}
\usepackage{soul}
\usepackage{csquotes}
\usepackage{microtype}      
\usepackage{lipsum}
\usepackage{csquotes}
\usepackage{tikz}
\let\labelindent\relax
\usepackage{enumitem}
\usepackage{algorithm}
\usepackage{algorithmic}
\usepackage{gensymb}
\usepackage[utf8]{inputenc}

\newcommand{\xxnote}[3]{}
\ifx\hidenotes\undefined
  \usepackage{color}
  \renewcommand{\xxnote}[3]{\color{#2}{#1: #3}}
\fi

\newcommand{\bbm}{\begin{bmatrix}}
\newcommand{\ebm}{\end{bmatrix}}

\graphicspath{{figs/}}

\title{\LARGE \bf
Can a Robot Become a Movie Director? \\Learning Artistic Principles for Aerial Cinematography
}

\author{Mirko Gschwindt$^{1}$, Efe Camci$^{2}$, Rogerio Bonatti$^{3}$, Wenshan Wang$^{3}$, Erdal Kayacan$^{4}$, Sebastian Scherer$^{3}$
\thanks{$^{1}$Mirko Gschwindt is with the Department of Computer Science, Technische Universität München (TUM), Munich, Germany.
        {\tt\small m.gschwindt@tum.de}}
\thanks{$^{2}$Efe Camci is with School of Mechanical and Aerospace Engineering (MAE), Nanyang Technological University (NTU), 50 Nanyang Avenue, 639798, Singapore.
        {\tt\small efe001@e.ntu.edu.sg}}
\thanks{$^{3}$Rogerio Bonatti, Wenshan Wang, and Sebastian Scherer are with The Robotics Institute, School of Computer Science, Carnegie Mellon University, Pittsburgh PA.
        {\tt\small \{rbonatti,wenshanw,basti\}@cs.cmu.edu}}
\thanks{$^{4}$Erdal Kayacan is with Department of Engineering, Aarhus University, DK-8000 Aarhus C, Denmark.
        {\tt\small erdal@eng.au.dk}}
}

\begin{document}

\maketitle
\thispagestyle{empty}
\pagestyle{empty}

\begin{abstract}
Aerial filming is constantly gaining importance due to the recent advances in drone technology. It invites many intriguing, unsolved problems at the intersection of aesthetical and scientific challenges. In this work, we propose a deep reinforcement learning agent which supervises motion planning of a filming drone by making desirable shot mode selections based on aesthetical values of video shots. Unlike most of the current state-of-the-art approaches that require explicit guidance by a human expert, our drone learns how to make favorable viewpoint selections by experience. We propose a learning scheme that exploits aesthetical features of retrospective shots in order to extract a desirable policy for better prospective shots. We train our agent in realistic AirSim simulations using both a hand-crafted reward function as well as reward from direct human input. We then deploy the same agent on a real DJI M210 drone in order to test the generalization capability of our approach to real world conditions. To evaluate the success of our approach in the end, we conduct a comprehensive user study in which participants rate the shot quality of our methods. Videos of the system in action can be seen at \href{https://youtu.be/qmVw6mfyEmw}{https://youtu.be/qmVw6mfyEmw}.
\end{abstract}

\section{Introduction}

\begin{figure}[t]
    \centering
    \includegraphics[width=0.90\columnwidth]{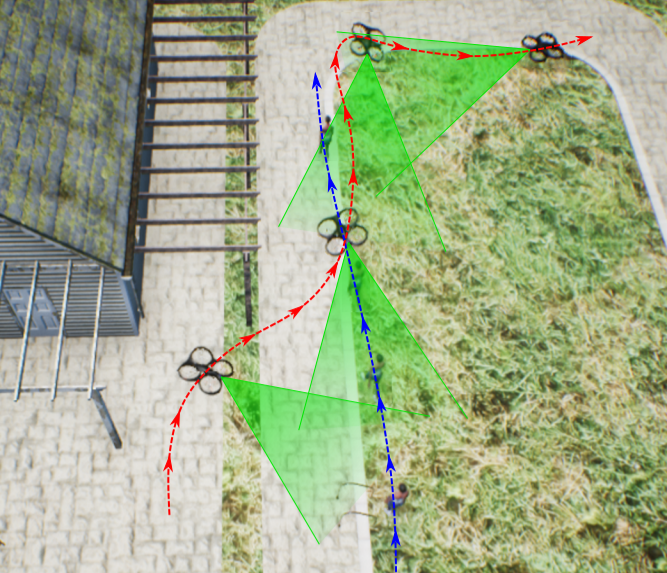}
    \caption{Time lapse of drone trajectory during filming in photo-realistic environment. Since the left hand side is occupied, the drone switches from left to front and then right viewpoint.}
    \label{trajectory_drone}
\end{figure}

\label{sec:introduction}
Aerial filming has invoked considerable attention within both large-scale drone companies and well-established research groups \cite{galvane2017automated,nageli2017real,galvane2018directing,gebhardt2018optimizing}. While the accessibility of personal filming drones is vast, there are still a number of research problems in this area: to create safer, more compact, aesthetically aware, user-friendly, and autonomous filming drones. In this work, we address aesthetical awareness of autonomous filming drones at the junction of well-known scientific problems such as motion planning of unmanned aerial vehicles in unknown environments and motion forecasting of moving targets, e.g., humans, cars, bicycles.

\begin{figure*}
\centering
\includegraphics[width=0.94\textwidth]{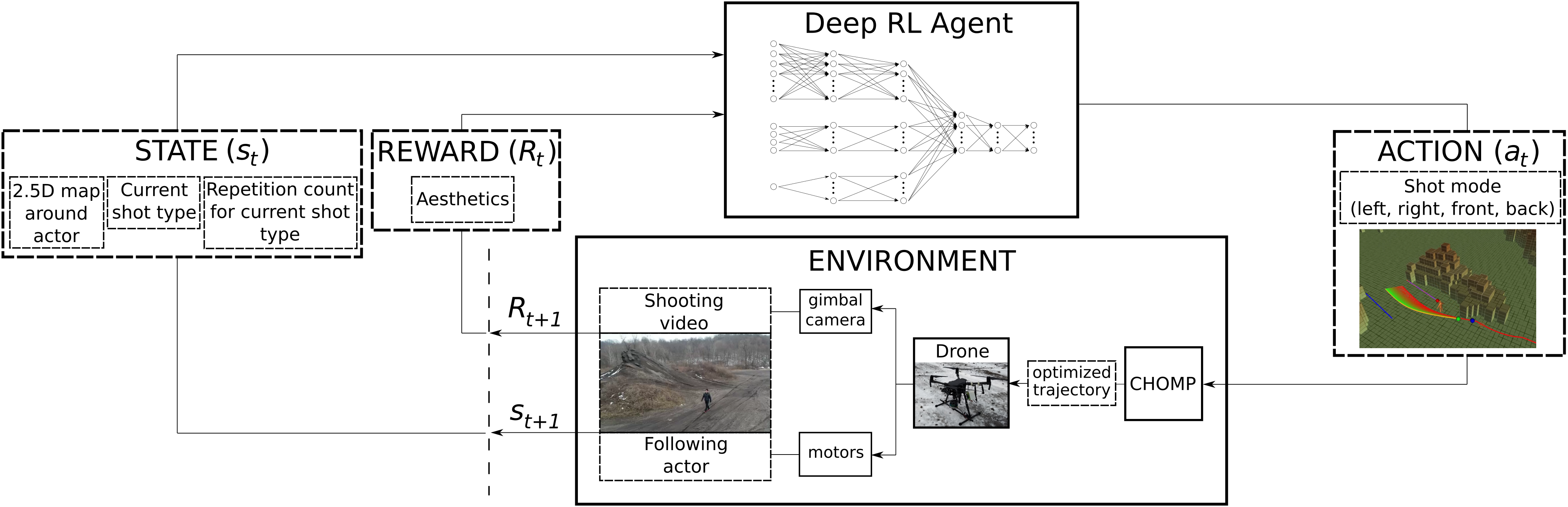}
\caption{Our overall system flow merged into a generic RL diagram.}
\label{system_diagram}
\end{figure*}

Current state-of-the-art methods for aerial filming require either complete or partial trajectory inputs from an expert user. In this case, the job of the expert pilot is arduous. The expert is required to solve a complex optimization problem intuitively in a short time frame by considering dynamic feasibility, collision avoidance, motion forecast of the actor, and visual aesthetics simultaneously. We attempt to automate this process by training a deep reinforcement learning (RL) agent with aesthetical awareness.

Deep RL has started revolutionizing many research areas in recent years. It has shown great success in accomplishing tasks which are substantially difficult to be hard-coded or planned in advance such as playing Atari games \cite{mnih2015human}, Go \cite{silver2017mastering} and learning dynamic motor skills for legged robots \cite{hwangbo2019learning} or recovery maneuvers for multi-rotors \cite{hwangbo2017control}. Besides these tasks where a well-specified reward function is available, it has yielded promising results when designing a suitable reward function is non-trivial such as socially compliant robot navigation \cite{kretzschmar2016socially,fu2017learning}. While both classes of problems are already challenging, task success subjectivity makes the learning process even harder in some cases. To solve these tasks deep RL has successfully incorporated human preferences (human reward functions) \cite{hadfield2016cooperative,christiano2017deep} while achieving a higher level goal. Inspired by the competitive success of these RL algorithms, we propose a novel RL system for autonomous aerial filming by incorporating both hand-crafted and human reward functions in this work. 

We build upon our previous work \cite{bonatti2018autonomous} in which smooth and safe trajectories are generated in real-time using CHOMP \cite{ratliff2009chomp}. In our previous motion planning pipeline, CHOMP is utilized to optimize the cinematographic trajectories by considering cost functions for smoothness, shot quality, safety, and occlusion avoidance. The process is detailed further in \cite{bonatti2019iros}. However, it needs a higher level input from the user, i.e., shot mode, which defines the raw cinematographic path. It uses this path as a baseline while conducting trajectory optimization. In order to automate the overall filming process even further, we train a deep RL agent which provides the shot type input intelligently, replacing the user input (Fig.~\ref{system_diagram}). Starting from no prior knowledge, the agent selects different shot types (action) for different situations (state) and observes their corresponding values (reward) based on certain aesthetical criteria. While these criteria are coded into a complex mathematical function considering widely-recognized cinematographic rules for the hand-crafted reward, they are solely humans' aesthetical assessment in the case of the human reward function. Using this retrospective knowledge enhanced over numerous trials, the agent explores a desirable policy which is able to replace external user inputs and supervise the whole filming process autonomously. The specific contributions of this work are:
\begin{itemize}
    \item A deep RL agent which supervises viewpoint selection for aerial filming;
    \item Incorporating human preferences into aesthetical metrics;
    \item A comprehensive user study to evaluate our method;
    \item Real world tests with a filming drone.
\end{itemize}


\section{Related Work} 
\label{sec:related_work}
Intelligent shot mode or viewpoint selection for aerial camera control borrows concepts from the disparate fields of virtual cinematography, aerial filming, human-preference learning, and learning artistic beauty. We go through each of these briefly in the next subsections.
\subsection{Arts and computer graphics}
Camera control in virtual cinematography has been extensively examined by the computer graphics community \cite{christie2008camera}. These methods usually employ through-the-lens control where a virtual camera is manipulated while maintaining focus on certain image features \cite{gleicher1992through,drucker1994intelligent,lino2011director,lino2015intuitive}. Artistic features for camera control follow different empirical principles known as composition rules \cite{arijon1976grammar,bowen2013grammar}.

Several works analyse the choice of which viewpoint to employ for a particular situation. For example, in
\cite{drucker1994intelligent}, the researchers use an A* planner to move a virtual camera in pre-computed indoor simulation scenarios to avoid collisions with obstacles in 2D. 
More recently, we find works such as \cite{leake2017computational} that 
post-processes videos of a scene taken from different angles
by automatically labeling features of different views. The approach uses high-level user-specified rules which exploits the formerly labeled features in order to automatically select the optimal sequence of viewpoints for the final movie. Besides, \cite{wu2018thinking} helps editors by defining a formal language of editing patterns for movies.


\subsection{Autonomous aerial cinematography}
We also see works specific to aerial cinematography using unmanned aerial vehicles. For example, \cite{gebhardt2018optimizing,roberts2016generating,joubert2015interactive,xie2018creating} focus on key-frame navigation, filming static landscapes or structures. \cite{lan2017xpose} also touches the idea of image key-frames, which are defined by a user using through-the-lens control instead of positions in the world frame.

Another line of work on aerial filming focuses on tracking moving targets in dynamic contexts. For example, \cite{galvane2017automated,nageli2017real,galvane2018directing,bonatti2018autonomous,joubert2016towards} present different approaches with varying levels of complexity in terms of obstacle and occlusion avoidance as well as real-life applicability.

\subsection{Making artistic choices autonomously}
A common theme behind all the work presented so far is that a user must always specify which kind of output they expect from the system in terms of artistic behavior. If one wishes to autonomously specify artistic choices, two main points are needed: a proper definition of a metric for artistic quality of a scene, and a decision-making agent which takes actions that maximize this quality metric.

Several works explore the idea of learning a beauty or artistic quality metric directly from data. \cite{karpathy2015deep} learns a measure for the quality of \textit{selfies}; \cite{fang2017creatism} learns how to generate professional landscape photographs; \cite{gatys2016image} learns how to transfer image styles from paintings to photographs. 

On the action generation side, we find works that have exploited deep RL \cite{mnih2015human} to train models that follow human-specified behaviors. Closely related to our work, \cite{christiano2017deep} learns behaviors for which hand-crafted rewards are hard to specify, but which humans find easy to evaluate. In the field of autonomous drone filming \cite{Huang_2019_CVPR} tries to learn different drone filming styles by deploying an imitation learning approach.

Our work, as described in Section~\ref{sec:approach}, brings together ideas from all the aforementioned areas to create a generative RL model for shot type selection in aerial filming drones which maximizes an artistic quality metric and specifically considers the surrounding environment.





%





\section{Approach} 
\label{sec:approach}

\subsection{Learning objectives}
The main aim of our deep RL agent is to supervise motion planning of a filming drone intelligently by selecting desirable viewpoints at the right time. The agent is expected to create desired viewpoint sequences while satisfying the following conditions:
\begin{itemize}
    \item The actor is in view and within desired shot angle limits.
    \item The overall video switches shot directions to not only show one side of the actor and keep the shot interesting.
    \item The drone does not collide with obstacles.
    \item The overall video sequence is aesthetically pleasing.
\end{itemize}
\subsection{RL Problem Formulation}
The main challenge in applying RL successfully in real-world scenarios is to formulate the problem in such a way that the agent is able to derive useful representations of the environment and it is able to exploit these to generalize to new situations using an evaluative feedback, i.e., reward. In this vein, the proposed approach can be examined in three folds, each referring to the main elements of RL, i.e., state, action, and reward.  
\subsubsection{State}
In the proposed approach, state ({\it s}) consists of three elements: a 2.5D local height map around the actor, current shot type, and repetition count for the current shot type. The 2.5D map around the actor gives insight into obstacles close to the actor and paths the actor is likely to take based on obstacle locations. It is an informative representation of the local environment which is composed of a grayscale image in which each pixel has a value governing the highest obstacle occupying that grid. Alongside its informativeness, it is compact enough to be fed into a deep Q-network (DQN) easily. It is represented as a 24$\times$24 matrix containing values between 0 and 255.

The current shot type is included in the state definition in order to provide a grasp of favorable switching between different modes. That is, although a particular shot type works the best for a given situation, direct switching to that mode from the agent's current shot mode may be undesirable due to a difficult transition between two shot modes considering the control effort for the drone. While the shot types available in today's filming drones have a variety such as left/right, front/back, orbit, high pan, fly-through, etc., we focus on the four basic types in this work: left, right, front, and back shot. The current shot type is coded as a one-hot vector in order to be fed to the DQN easily.

The repetition count for the current shot type is a number that states for how many time steps the current shot type has been applied repeatedly since the last switch, where one time step, and therefore the time between each drone decision, lasts 6s\footnote{Time step duration is selected as 6s due to a limitation enforced by our low-level planning algorithm, CHOMP. When a new shot mode is selected on-the-fly, it takes time for CHOMP to generate an intermediate path which will move the drone from its current viewpoint to the newly selected viewpoint. The duration of 6s has been observed to be a decent time interval for this purpose.}. Because keeping the same shot type for a long duration would possibly be undesirable from a cinematographic perspective, this number provides insight into whether the shot stagnates so long that it might become boring. This number is normalized to a value between 0 and 1 before being fed to the DQN.

\subsubsection{Action}
The action is the shot mode that the agent selects at the beginning of each time step. Similar to the current shot type element of state, our action space is composed of four different shot modes, i.e., left, right, front, and back. Each shot mode selection results in a new desired cinematographic path for our low-level planning algorithm (CHOMP) at each time step. The essential aim of the agent is to explore the desirable shot mode sequences in a given situation to maximize the sum of the rewards.

\subsubsection{Hand-crafted reward}

We attempt to encode the artistic requirements with four elements:

\textit{Shot angle} $R_{sa}$ is related to the current UAV tilt angle $\theta_t$ relative to the actor. We define an optimal value $\theta_{t,opt}$ and an accepted tolerance $\theta_{t,tol}$ around it. Between those we decay the reward linearly from $R_{sa} = 1.0$  at $\theta_t=\theta_{t,opt}$ to $R_{sa} = 0.0$ at the tolerance boarders $\theta_t=\theta_{t,opt} \pm \theta_{t,tol}$. We assign a negative punishment of $R_{sa} = -0.5$, if the angle falls out of bounds.

\textit{Actor's presence ratio} $R_{pr}$ is related to the space that the actor occupies in the camera image compared to the image size. Based on the shot scale, actor size, and camera parameters, we set two bounds $pr_{min}$ and $pr_{max}$ and use the heuristic that an ideal presence ratio lies in between them\footnote{The boundaries for shot angle and presence ratio are a result of empirical data gained from numerous informal trials.}. If the detected presence ratio lies inside these bounds we pass on our previous reward $R_{pr}=R_{sa}$. Otherwise, we assign a negative punishment of $R_{pr} = -0.5$.

The relative UAV tilt angle and the actor's presence ratio are measured for each frame over a video clip ($N$ frames in a video clip) obtained during each time step and the resulting rewards are averaged to an intermediate reward $R_{avg}$ at the end of each time step:
\begin{equation}
\label{eq:reward_average}
    R_{avg} = \frac{1}{N}\sum_{i=1}^N R_{pr,i}
\end{equation}

\begin{figure}[h]
\centering
\includegraphics[width=0.7\columnwidth]{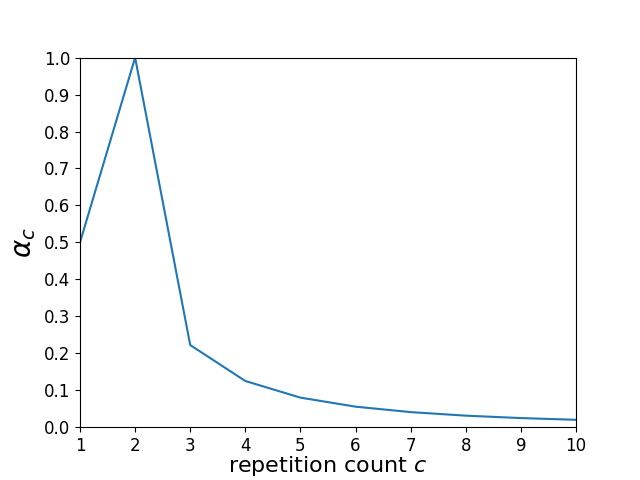}
\caption{Adaptive reward coefficient $\alpha_c$ over shot repetition count $c$, assuming an optimal repetition count $c_{opt}$ of 2.}
\label{fig:reward_discount}
\end{figure}

\textit{Shot type duration:} $\alpha_c$ is a reward coefficient related to the length of the current shot type given by a repetition count $c$. We use the heuristic that an ideal shot length is of $c_{opt}$ time steps, and define $\alpha_c$ in Equation~\ref{eq:discount_shot_counter}. Assuming $c_{opt}=2$ for example, $\alpha_c$ varies over $c$ as in Fig.~\ref{fig:reward_discount}.

\begin{equation} \label{eq:discount_shot_counter}
\alpha_{c} = 
        \begin{cases}
            \frac{c}{c_{opt}},& \text{if } c\leq c_{opt}\\
            \frac{c_{opt}}{c^2},& \text{otherwise}
        \end{cases}
\end{equation}

We discount our reward for shot repetition as follows:

\begin{equation} \label{eq:reward_discounted}
R_{discounted} = \begin{cases}
             R_{avg} \cdot \alpha_{c} ,              & \text{if } R_{avg}\geq 0\\
            \frac{R_{avg}}{\alpha_{c}},              & \text{otherwise.}
        \end{cases}
\end{equation}

\textit{Collision punishment} $R_{cp}$ is related to UAV safety. If the UAV crashes during a time step, $R_{cp} = -1.0$; otherwise, $R_{cp} = 0$.

We define the total reward $R$ for a time step as:

\begin{equation} \label{eq:reward_final}
R = \begin{cases}
             R_{cp},              & \text{if } R_{cp} = -1.0\\
             R_{discounted},      & \text{otherwise.}
        \end{cases}
\end{equation}

\subsubsection{Reward from human preferences}
\begin{figure}[t!]
\centering
\includegraphics[width=0.9\columnwidth]{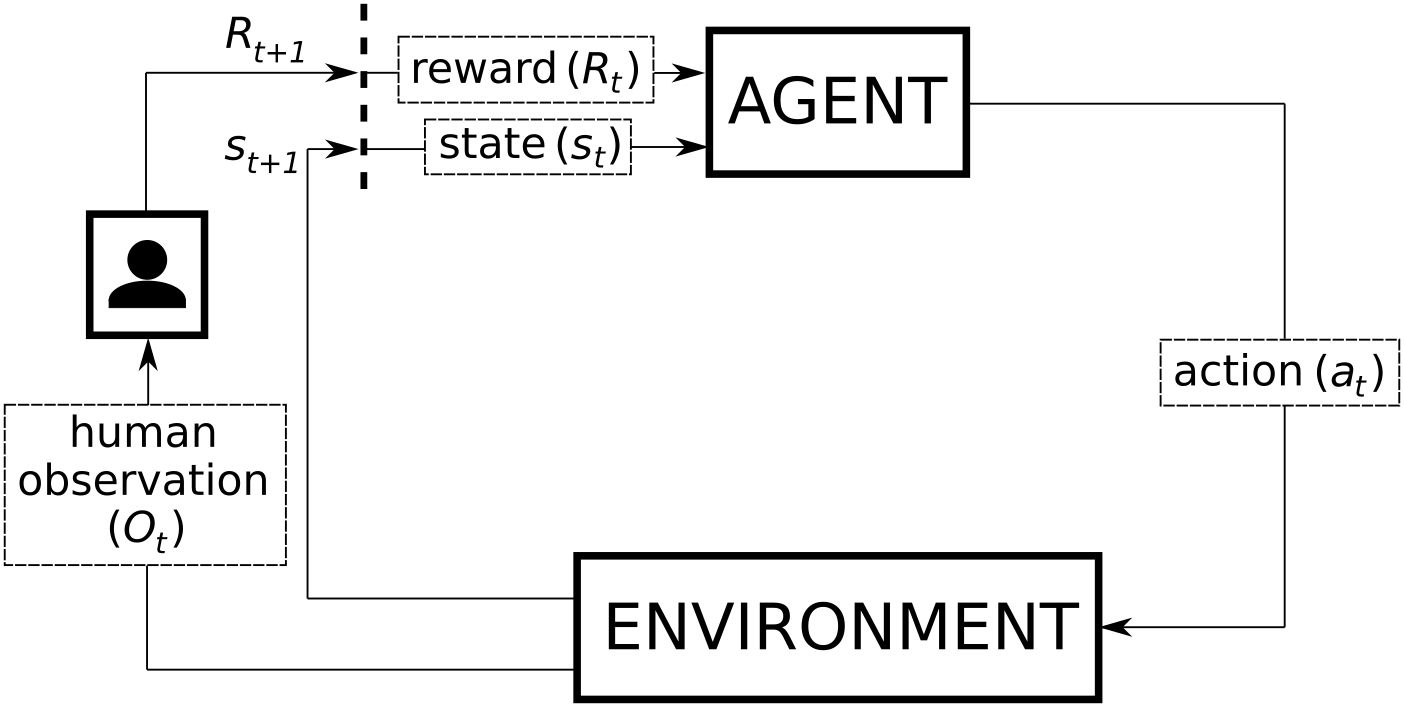}
\caption{Training pipeline with human in the loop.} 
\label{human_reward}
\end{figure}

Because it can be difficult to define aesthetical values in a mathematical reward function precisely, we test another training approach where the reward is directly generated by the human perception of aesthetics (Fig. \ref{human_reward}). We use the same training pipeline with the same hyperparameters for our human reward training as for all our other training sessions. The only difference being that the reward given during training is not delivered by a mathematical function, but by humans rating the shot quality of the scene on-the-fly. During training, multiple people from our lab constantly rate the observed camera image. Every time step (6s), the rating person is asked to give an evaluation from 1 to 5 stars for shot quality, or 0 for a collision. This rating is mapped linearly to a reward ranging from -0.5 to +1, and -1 for a collision, respectively. The range of rewards resembles the possible rewards from our hand-crafted reward configuration.

\subsection{Algorithmic Details}
The essential idea behind the use of DQN is to estimate the highly complex action-value function $Q(s,a)$ which yields the value of an action taken on a particular state considering its long-run gains for the agent. It is formally defined as: 
\begin{equation} \label{eq_qfunction}
Q(s,a)=\mathbb{E} \left(R_{t+1}+\gamma \displaystyle \mathop{\max}_{a'} Q(s_{t+1},a') \big| s_t=s, a_t=a \right)
\end{equation}
where $s_t$ is the current state and $a_t$ is the action taken. The terms $s_{t+1}$ and $R_{t+1}$ are the next state that the agent reaches and the reward it observes, respectively. The term $\gamma$ is the discount factor which determines the present value of future rewards \cite{sutton1998reinforcement}. 

\begin{figure}[t!]
\centering
\includegraphics[width=\columnwidth]{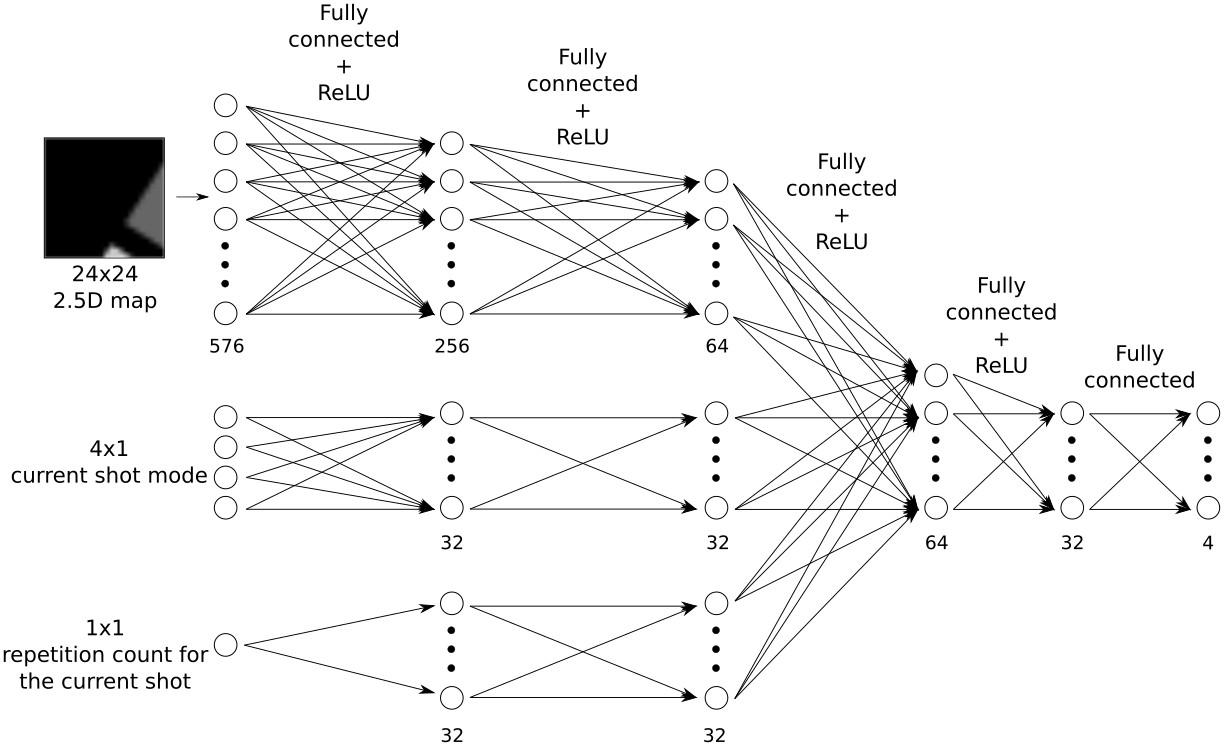}
\caption{DQN architecture.}
\label{dqn_arch}
\end{figure}

The DQN considered in this work is composed of fully connected neural networks which fuse three different inputs in three main lanes. The architecture is depicted in detail in Fig. \ref{dqn_arch}. Three inputs (2.5D local map, current shot type, repetition count for the current shot type) are first fed through three different lanes separately. Then, they are combined in a single lane which eventually yields the Q values for each action. All consecutive layers are fully connected to each other, and all of them apply ReLU (rectified linear unit) with the exception of the very last layer. We train this network using Adam optimizer \cite{kingma2014adam} with default settings in PyTorch. We use Huber loss \cite{huber1964robust} for creating the gradient for the parameter update.

The pseudocode for overall training using experience replay (ER) is given in Algorithm \ref{algo}.

\begin{algorithm}[t!]
\caption{Pseudocode for training DQN.}
\label{algo}
\begin{algorithmic}
\FOR{i \textbf{in} episodes}
\STATE{observe state $s$}
\FOR{t \textbf{in} timesteps}
\STATE{take action $a$ governed by $\varepsilon$-greedy policy} 
\STATE{observe state $s'$ and reward $R$}
\STATE{save data sample ($s$,$a$,$s'$,$R$)}
\STATE{$s$ := $s'$}
\STATE{update experience replay (ER) with the new data}
\ENDFOR
\IF{i $\%$ min episodes for update == 0}
\STATE{divide ER into random minibatches of size $n$}
\FOR{j \textbf{in} minibatches}
\STATE{apply $y_{1:n}$ = $R_{1:n}$ + $\gamma \mathop{\max}_{a'} Q$(${s'}_{1:n}$,$a'$; $\theta_i$)}
\STATE{update DQN weights by $L_{\delta}$($y_{1:n}$ - $Q$($s_{1:n}$,$a_{1:n}$; $\theta_i$))}
\ENDFOR
\ENDIF
\ENDFOR
\end{algorithmic}
\end{algorithm}
\section{System} 
\label{sec:system}

\subsection{Simulation}
We train our agent exclusively in simulation. The environment we use for training is the Microsoft AirSim (\cite{airsim2017fsr}). AirSim is an open source simulation environment for drones and cars that is based on the games engine Unreal Engine. It offers the ability to simulate highly realistic conditions, especially for aerial vehicles.

\subsubsection{Artificial Worlds}
\begin{figure}
    \centering
    \includegraphics[width=0.95\columnwidth]{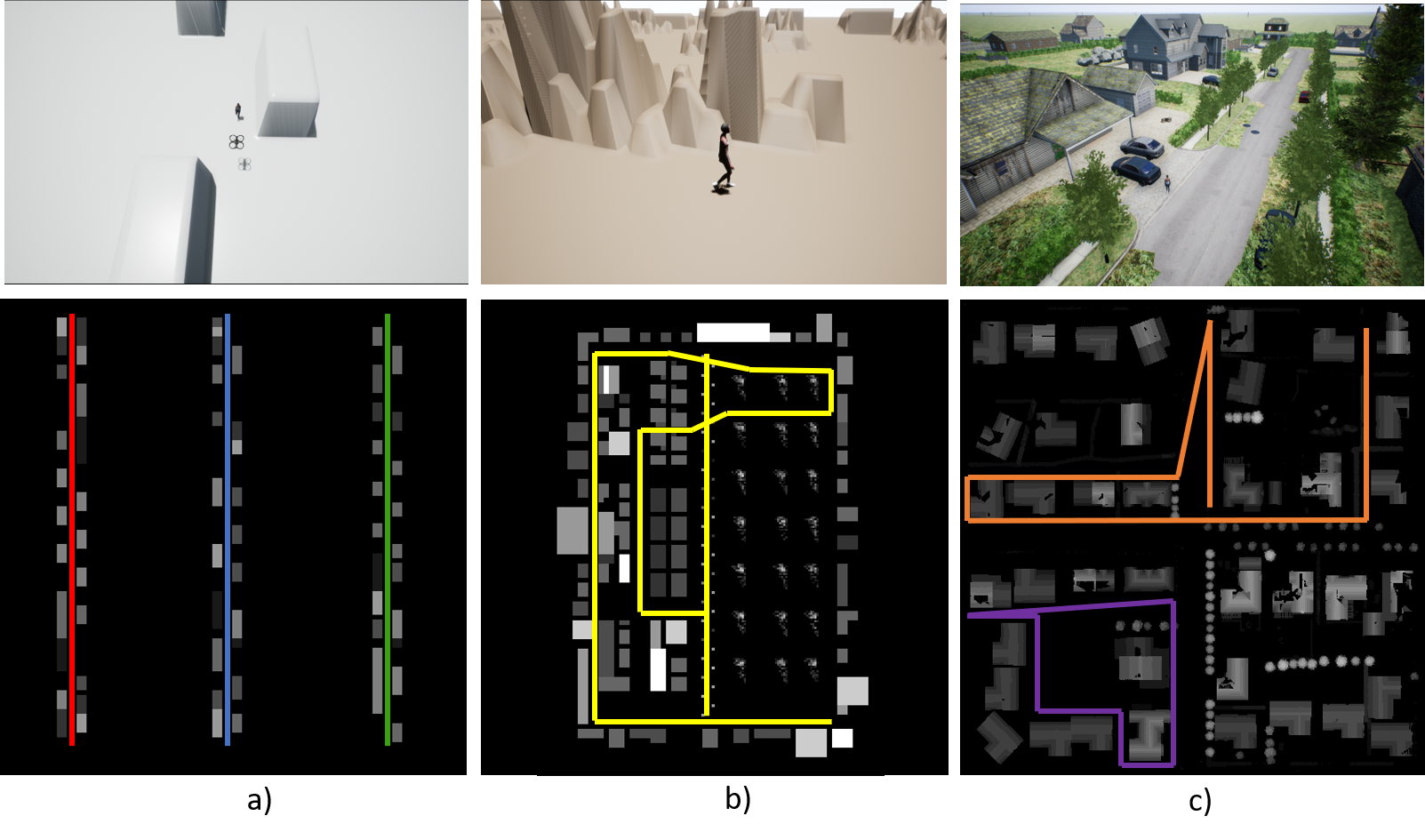}
    \caption{Screenshots and height maps of AirSim environments. Actor routes: a) red: block world training route, blue: block world test route 1, green: block world test route 2, b) yellow: bigmap long route, c) orange: neighborhood training route, purple: neighborhood test route.}
    \label{heightmaps}
\end{figure}

We train our agent in different types of artificial world settings. In our simple block environment, the actor is walking on a path with alternating blocks on the left and right side in varying heights and lengths (Fig. \ref{heightmaps}a).

Our second artificial training environment is a much larger map that is separated into three distinguishable areas: a slightly more complex block area, an area with two parallel rows of pillars, and an area with mountain-like structures (Fig. \ref{heightmaps}b). In contrast to the former simple block environment, the block section of the larger map features varying sizes of corridors, more different block shapes, and better possibilities for traversing in multiple directions. This section is created to resemble a city landscape.

The column section features two rows of lean pillars in varying heights. In contrast to the block section, the obstacle shapes are much slimmer. 

The mountain section features three rows of mountain-shaped structures. All mountains are similar in shape and modeled after a real world drone testing environment near Pittsburgh. The exact height and shape of the mountains are randomly created within a predefined range. 

There are four training routes that the actor can walk along in the larger environment: through the block section, through the pillars, in between the mountains, and a large route through the whole map. We also let the actor walk randomly over the map (`roaming'). In this mode, the actor randomly selects a point on the map and takes the shortest route towards that point. Afterwards, the actor randomly chooses the next point.

\subsubsection{Photo-realistic World}
To showcase scenarios closer to real world environments, we also train our agent in a slightly modified version of the `Neighborhood' environment by Unreal Engine (Fig. \ref{heightmaps}c). This environment offers a photo-realistic simulation of a suburban residential area that could act as the place of a movie scene. We create an actor trajectory along streets, houses, bushes, trees, and cars to have a variety of obstacles that could interfere with the drone. We also make use of the roaming approach for training again.

\begin{figure}[t!]
    \centering
    \begin{minipage}{0.49\columnwidth}
    \centering
    \includegraphics[height=0.12\textheight]{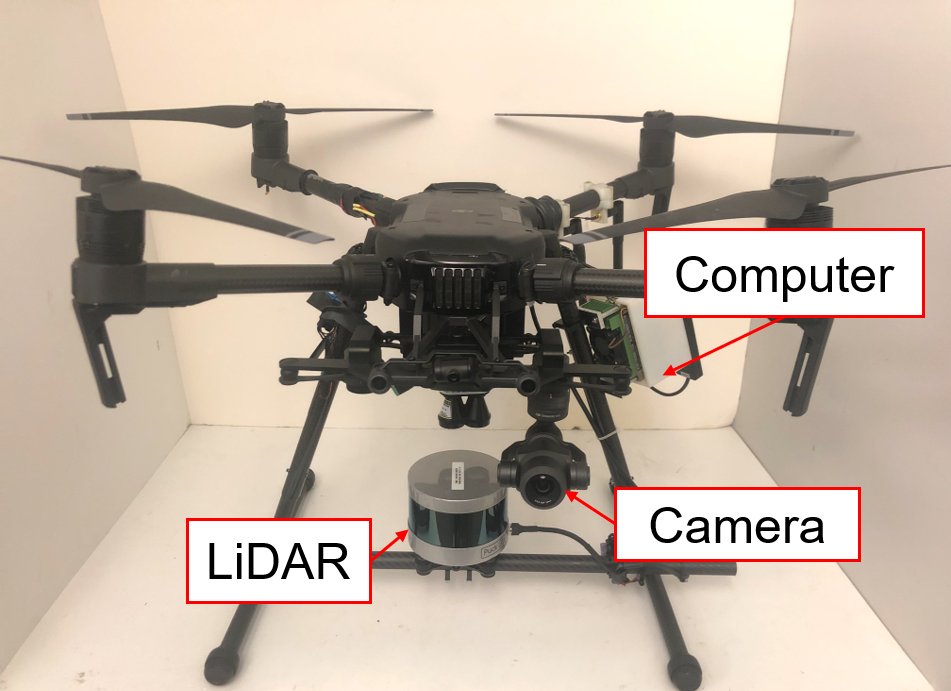}
    \end{minipage}
    \centering
    \begin{minipage}{0.49\columnwidth}
    \centering
    \includegraphics[height=0.12\textheight]{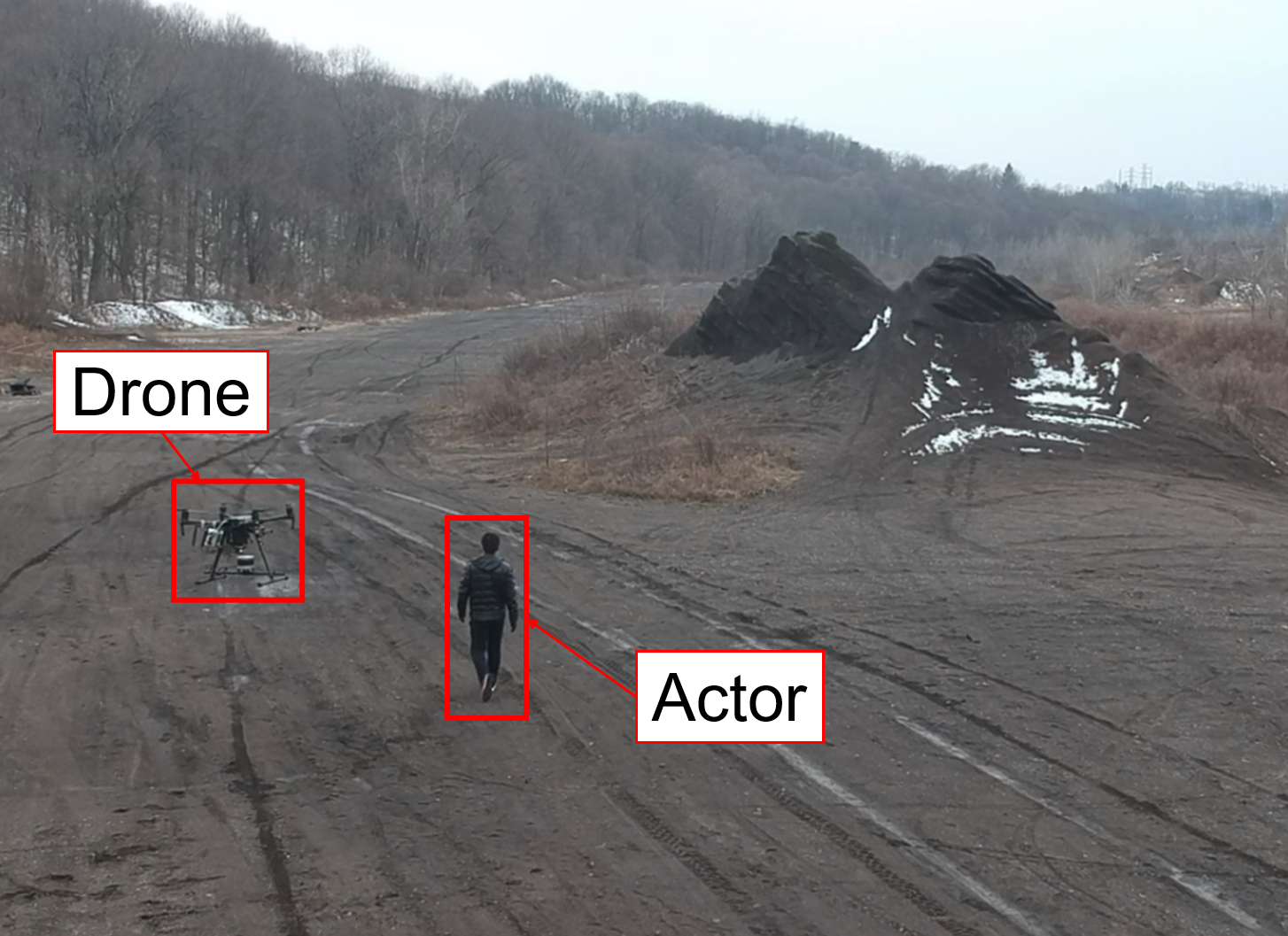}
    \end{minipage}
    \caption{Real robot with its on-board sensors on the left and in action in the test facility on the right (drone marked by red square).}
    \label{drone}
\end{figure}

\subsection{Hardware}
We train our agent exclusively in simulation. However, for testing, we deploy our algorithm in real world conditions. Our platform is a DJI M210 drone, shown in Fig. \ref{drone}. All processing is done with an 
NVIDIA Jetson TX2 computer, with 8GB of RAM and 6 CPU cores. An independently controlled gimbal DJI Zenmuse X4S records high-resolution images. We operate in pre-mapped environments, for which we generate a height map that is later  cropped locally and used as one of the inputs for the network. 

Our pre-mapped testing facility near Pittsburgh, PA has a heap of rubble that is used as the obstacle for the experiments.



\section{Results} 
\label{sec:results}
For the purpose of presenting training and testing results, we consider 5 consecutive time steps (each with a duration of 6s) per episode. Therefore, each episode entails 5 decision points and 30 seconds of filming. Training is performed over 300-2000 episodes, depending on the complexity of the environment and the length of the training route.
\subsection{Hand-crafted reward}
\subsubsection{Artificial worlds}
\begin{figure}
    \centering
    \begin{minipage}{0.49\columnwidth}
    \centering
    \includegraphics[width=\linewidth]{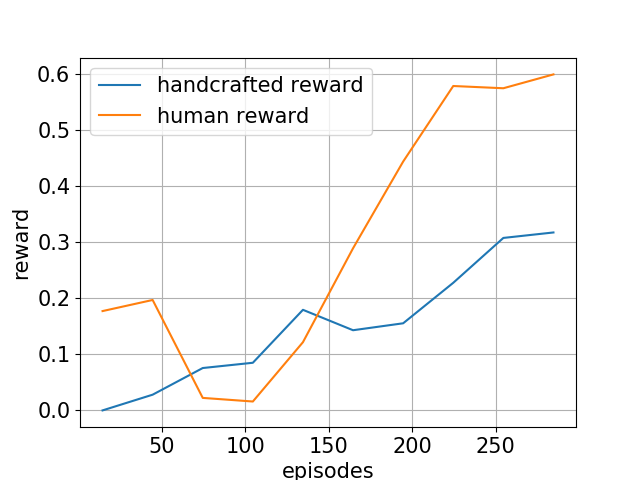}
    \end{minipage}
    \centering
    \begin{minipage}{0.49\columnwidth}
    \centering
    \includegraphics[width=\linewidth]{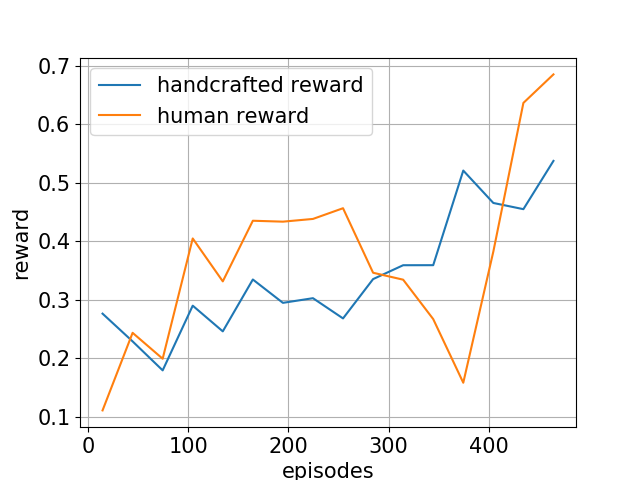}
    \end{minipage}
    \caption{Reward curves during training for hand-crafted reward and human reward (30 episodes grouped to one data point). left: block world, right: neighborhood.}
    \label{plot_reward_curve}
\end{figure}

We train the agent in our simple block world for 300 episodes (Fig. \ref{plot_reward_curve}). This environment offers a decent opportunity to evaluate the performance of the trained agent as it has a clear pattern of blocks that the drone can avoid. Our agent performs significantly better than a random policy in the training environment, as well as in the two previously unknown testing environments (Table~\ref{testing_blockworld})\footnote{The standard deviation for rewards of all trained policies is in the magnitude of 0.1.}. Our agent is able to achieve all goals that it was supposed to learn:
\begin{itemize}
    \setlength\itemsep{0.1pt}
    \item It keeps the actor in view by preferring 90$\degree$ shot mode switches (i.e., from left to back or front) to 180$\degree$ switches (i.e., from left to right), which often lose sight of the actor.
    \item It switches shot types regularly to keep the shot interesting.
    \item It avoids flying above obstacles to keep the shot angle in desirable limits.
    \item It avoids extremely high obstacles which might pose a threat of crashing.
\end{itemize}
It achieves these goals both in the training map and in the testing maps.

\begin{table}
\centering
\caption{Average rewards per time step for testing in block world environment.}
\begin{tabular}{l | c c c}
\hline
 & \textbf{Training map} & \textbf{Test map 1} & \textbf{Test map 2} \\
\hline
\textbf{Random policy} & -0.0061 & 0.0616 & 0.0308 \\
\textbf{Trained policy} & 0.3444 & 0.3581 & 0.3662  \\
\hline
\end{tabular}
\label{testing_blockworld}
\end{table}

In our second artificial training environment (bigmap), we train our agent in different sections of the map in a range of 1000-2000 episodes (Table~\ref{testing_bigmap}). In all routes, a significant increase of rewards over a random policy can be noticed. Even around more complex structures like mountains, our agent satisfies our goals of keeping the actor in view, keeping the shot within the desired shot angle limits, switching shot type for an interesting scene, and avoiding high obstacles due to the danger of crashes. Only in the pillar section of the environment, the trained policy is not able to avoid occlusions. This is a result of our reward function not punishing short occlusions very harshly. Due to the thin, cylindrical shape of the pillars, the agent will only be occluded for very short time frames, thus the rewards are still acceptable.

\subsubsection{Photo-realistic world}
In our photo-realistic neighborhood environment, we train our agent on a fixed training route for 500 episodes (Fig.~\ref{plot_reward_curve}) and in roaming mode for 2000 episodes.

\begin{table}[]
    \centering
    \caption{Average rewards per time step for policies trained in bigmap environment. All policies are tested on their respective training route.}
    \begin{tabularx}{.85\columnwidth}{l|c c c c}
    \hline
         & \textbf{Mountain} & \textbf{Block} & \textbf{Pillar} & \textbf{Long}\\
         & \textbf{route} & \textbf{route} & \textbf{route} & \textbf{route}\\
         \hline
        \textbf{Random policy} & 0.2654 & 0.1203 & 0.2319 & 0.1944\\
        \textbf{Trained policy} & 0.5908 & 0.3957 & 0.4687 & 0.5052\\
        \hline
    \end{tabularx}
    \label{testing_bigmap}
\end{table}

\begin{table}
\centering
\caption{Average rewards per time step for testing in neighborhood environment.}
\begin{tabularx}{0.95\columnwidth}{l | c }
\hline
 & \textbf{Neighborhood test route}\\
\hline
\textbf{Random policy}  & 0.2047  \\
\textbf{Bigmap long route policy} & 0.5760 \\
\textbf{Bigmap block section policy} & 0.4638\\
\textbf{Neighborhood training route policy} & 0.5882\\
\textbf{Neighborhood roaming policy} & 0.5546 \\
\hline
\end{tabularx}
\label{testing_neighborhood}
\end{table}

Our results in the photo-realistic environment (Table~\ref{testing_neighborhood}, Fig. \ref{trajectory_drone}) show that our agent is capable of handling real world scenarios. While the policy trained on the training route of the neighborhood environment performs best, the policy trained on the long route in the bigmap environment gets surprisingly close, confirming a successful generalization from an artificial map to a realistic environment. The policy trained solely in the block section of bigmap performs significantly worse, despite the idea of the blocks resembling the shapes of houses. It seems like the policy overfits to the block shape and cannot handle different shapes like trees and houses with sloping roofs.

\subsubsection{Real world}
To test our trained policy in a real world setting, we implement our shot selection algorithm on the aforementioned DJI M210 drone. We test the algorithm around a heap of rubble in the Gascola region near Pittsburgh, PA (Fig. \ref{drone}). For the testing procedure, we deploy a DQN which was previously trained in the mountain section of the bigmap environment. We shoot 3 video clips comparing 3 different drone filming policies:
\begin{itemize}
    \item our trained policy;
    \item a policy that stays exclusively in the back of the actor;
    \item a policy that selects an action randomly.
\end{itemize}
Our trained policy is able to transfer the learned objectives to the real world environment. It chooses shot modes based on the environment, the heading of the actor, and the current drone position relative to the actor. The shot selection follows all previously learned principles to form a smooth, occlusion-free, and interesting film scene. Our algorithm performs in real time. A forward pass through the network to select the next shot takes 10ms.

The random policy and the back shot policy serve as a reference for the shot quality. Neither of these policies result in a satisfactory video clip. The back shot policy produces a very stable, but also unexciting shot while the random policy results in a lot of turbulent drone movements that lose sight of the actor multiple times. Only the trained policy is able to produce a satisfactory video clip.

\subsection{Reward from human preferences}
Visual aesthetics are subjective and difficult to define through a mathematical function. Therefore, we compare the results of our hand-crafted reward policy to the  results of a policy trained through human reward. Since the amount of time and effort which the participants can dedicate are limited, this training is only performed in two maps, the simple block world and the neighborhood environment for 300 episodes and 500 episodes, respectively (Fig.~\ref{plot_reward_curve}). The valleys in the reward curves are caused by different human raters having different subjective evaluations of the quality of a shot. While the reward curve is not as steadily increasing as in the previous training sessions, the final reward and the difference to the initial reward are a lot higher, indicating a strong increase in knowledge. To get a numerical evaluation of the human trained policies and compare them to our previous policies, we conduct a user study.

\subsection{User study results}
In the user study for evaluating drone filming policies, we ask 10 participants to watch video clips taken by different policies and to order them as well as to write a short comment about each clip. We use 4 different policies to create the video clips:
\begin{itemize}
    \item our policy trained using the hand-crafted reward function;
    \item our policy trained by human preferences;
    \item a policy that always stays in the back of the actor;
    \item a policy that selects actions at random.
\end{itemize}
This way, we can compare both of our own policies to each other as well as to two baseline policies. The back shot policy takes the role of the safe option that can film the actor in almost any situation without having occlusion problems. The random policy is there as a sanity check to ensure that our trained policies are performing better than arbitrary actions. \\
During the study, we present participants with a total of 5 scenes, two from the simple block world environment and three from the neighborhood environment. The scenes are selected based on interesting obstacles. None of the scenes were previously encountered by either of the trained policies. For the simple block world environment, the policies are trained for 300 episodes in the training map and the video clips taken in the first testing map. For the neighborhood environment scenes, we train our agent for 500 episodes on the neighborhood training route. The scenes are chosen from different parts of the map not featured on the training route.

For each of the scenes, participants watch four 30s-video clips, one for each policy. The order of the video clips is randomized. After each scene, participants are asked to order the clips from `most visually pleasing' to `least visually pleasing' and to write a short comment for each clip.

The results of the user study are shown in Table~\ref{userstudy_results}. 

\begin{table}
\caption{Average score of video clips in the user study (from 0: worst to 10: best) via linear transformation from average rated position.}
\centering
\setlength{\tabcolsep}{0.5\tabcolsep}
\begin{tabularx}{\columnwidth}{l|cccccc}
\hline
 & \textbf{Average} & \textbf{Scene 1} & \textbf{Scene 2} & \textbf{Scene 3} & \textbf{Scene 4} & \textbf{Scene 5}\\
\hline
\textbf{Hand-crafted} & \textbf{8.2} & \textbf{10.0} & 5.3 & \textbf{9.3} & \textbf{7.7} & \textbf{8.7} \\
\textbf{Human} & 7.1 & 5.0 & \textbf{9.0} & 6.0 & \textbf{7.7} & 8.0 \\
\textbf{Back shot} & 3.8 & 4.0 & 4.7 & 4.3 & 4.0 & 2.0\\
\textbf{Random} & 0.9 & 1.0 & 1.0 & 0.3 & 0.7 & 1.3\\
\hline
\end{tabularx}
\label{userstudy_results}
\end{table}
\section{Discussion} 
\label{sec:discussion}

Our user study showcases that both our trained policies, one with a hand-crafted reward function and one trained through rewards via human input, perform significantly better than a random policy or one that only stays in one shot mode. Video clips from both policies are consistently rated the two highest in different environments on sceneries that were previously unknown to them. While the hand-crafted reward policy performs slightly better on average, participants' opinions are very divided in some cases.

The stated comments offer insight into the ranking of the four policies and human evaluation of drone filming aesthetics:
\begin{itemize}
    \item All participants criticize the back shot policy as `boring' or `unexciting'.
    \item All participants mention that the random policy loses view of the actor too often.
    \item The most common complaint is that the actor gets out of view. 
    \item There is a small window of how often the view angle should change. Many participants complain about too few changes, when only one shot angle change is performed. On the other hand, many participants criticize too much camera movement in a scene where the drone switches every time step.
    \item Participants frequently mention that they would like to get an overview of the surrounding instead of just seeing the actor in front of a building or a wall. When the drone gives multiple fields of view around the actor by switching the shot angle, this is positively remarked.
    \item The hand-crafted reward policy is often described as the most exciting while the human reward policy is described as very smooth.
\end{itemize}
The main difference between the hand-crafted reward policy and the human reward policy is the consistency of switching shots. Our hand-crafted reward function leads to a policy that tries to switch exactly every 2 time steps (12s), if not disturbed by an obstacle. Based on the user study results, this seems to be a relatively favorable time frame that achieves not being too boring while still giving enough time to get a good impression of the scene. The human reward policy is less consistent and switches the shot type more irregularly. We assume that an optimal switching frequency depending on the surrounding environment can be learned by more human training data input. Another interesting result of the study is that losing the view of the actor, even for a very short moment, is always rated as being very poor by the participants of the study. No policy that loses sight of the actor at any point is rated the highest in any scene. This leads to the conclusion to punish a loss of view of the actor even more harshly for future training episodes.
\section{Conclusion}
\label{sec:conclusion}
In this work, we present a fully autonomous drone cinematographer that follows a moving actor while making intelligent decisions about the shot type in real time. These decisions are based on previous experience, gained during training via deep RL with a hand-crafted as well as a human-generated reward function. Our approach works robustly in realistic simulation environments as well as in real world tests on a physical drone and successfully generalizes to previously unseen environments. The decisions about the shot direction acknowledge the environment around the actor and follow cinematographic principles such as occlusion avoidance, flat shot angles and frequent camera angle switches. Our user study confirms that our trained policies satisfy the human sense for aesthetics and offers insight into possible future improvements of the algorithm.

\section*{Acknowledgment}
This work was supported by Yamaha Motor Co., Ltd., Singapore Ministry of Education (RG185/17), and Aarhus University, Department of Engineering (28173). We also thank Cherie Ho, Xiangwei Wang, and Greg Armstrong for the assistance in field experiments and robot construction.

\bibliographystyle{IEEEtran}
\bibliography{IEEEexample}

\end{document}